\documentclass[11pt]{article}

\usepackage{acl}

\usepackage{times}
\usepackage{latexsym}
\usepackage{enumitem}

\usepackage{url}
\usepackage{subfigure}
\usepackage[T1]{fontenc}
\usepackage{tabularx}
\usepackage[utf8]{inputenc}
\usepackage{amsmath}
\usepackage{subcaption}
\usepackage{bbding}

\usepackage[T1]{fontenc}

\usepackage[utf8]{inputenc}

\usepackage{microtype}

\usepackage{inconsolata}

\usepackage{graphicx}

\title{Author-Informed Stance Detection: A Dataset and Findings}

\author{Ogen Schlachet Drukerman \and Einat Minkov\\
Faculty of Computer 
\& Information Science \\
University of Haifa \\}

\begin{document}
\maketitle
\begin{abstract}
Stance detection in social media is challenging, as interpreting a text often requires understanding the author’s perspective. We re-frame political stance detection as an author-informed text understanding task, where it is desired to infer both the author's political leaning and their stance towards the target from the text. We first show that human annotators struggle to infer authors’ political leanings and stances from individual tweets, and that pretrained large language models (LLMs) perform below human level in this setting. We then demonstrate that author context improves stance detection for both humans and LLMs, with particularly pronounced benefits for tweets perceived as sarcastic. Through extensive experiments, we further show that LLMs can be effectively fine-tuned to predict authors’ political affiliations, and that author-informed fine-tuning yields better stance detection performance than standard fine-tuning. This approach not only surpasses human performance but also improves generalization across datasets. To support further research on author-informed political stance detection, we release a large dataset of stance-labeled tweets enriched with author-context information.\footnote{Contact the authors for the link.} \textcolor{red}{Warning: Readers might find some of the examples offensive.}
\end{abstract}

\section{Introduction}

Stance detection in social media is a well-established NLP task with applications ranging from opinion polling to hate speech detection~\cite{aldayelIPM21}. It is commonly formulated as a classification task in which a text is labeled as {\it in favor of}, {\it against}, or {\it neutral} toward a given target~\cite{semeval16}. Despite extensive research, stance detection remains challenging. Although large pretrained language models (LLMs) have advanced the state of the art across many NLP tasks, their performance on stance detection often remains limited without supervised, in-domain fine-tuning~\cite{allawayEMNLP20,dingACL25}. However, such fine-tuning is resource-intensive and inherently subjective, as annotators may interpret the same text differently, often resulting in low inter-annotator agreement~\cite{sandriEACL23,belayTACL26}.

A key challenge for stance annotation and detection from text is that human communication typically occurs within a shared context. Interlocutors often have background knowledge of the topics being discussed and some familiarity with one another. This shared context allows speakers to express opinions indirectly, for example, through rhetorical devices such as sarcasm~\cite{operaACL19}, while still being understood by their intended audience~\cite{maynardEACL14}. Accordingly, recent research highlights the need for language agents to develop social awareness for understanding subjective language~\cite{yangCL25}, emphasizing the importance of recognizing the author's or speaker's perspective~\cite{flekACL20,mangalikACL25}.

In this work, we investigate stance detection in political tweets, with a particular focus on the role of user context. We posit that modeling an author's political perspective is an important component of political stance detection, as the identity of the stance-taker is central to stance-taking~\cite{DuBois2007} and can provide essential pragmatic context for interpreting social media discourse~\cite{pujariEMNLP24}. Consider, for instance, the tweet {\it ``Philly needs to elect more democrats, that should fix things.''} Based on the text alone, the tweet appears to express a positive stance toward {\it Democrats}. However, if the author is known to hold conservative political views, the statement may instead be interpreted as sarcasm, and the stance as negative. 
Nevertheless, an author's political orientation alone cannot determine their stance toward a specific target. Republican-affiliated users may criticize Trump, just as Democratic users may criticize members of their own political camp. Author context should therefore be treated not as a direct indicator of stance, but as a complementary source of information that can help disambiguate the author's intended meaning.

Existing research and datasets generally focus either on the stance expressed in text~\cite{semeval16,liACL21,kawintiranonACL21} or on the author's political orientation~\cite{shenEACL21,sucuACL25}, without explicitly integrating the two. 
To support {\it author-informed} stance detection, we introduce a dataset of more than 10K political tweets accompanied by information about the political affiliations of the authors, inferred from their social networks with high confidence~\cite{lotanPLOS23}. We conduct an extensive human annotation study to examine how well annotators can infer authors' political affiliations and identify tweet-level stances when given only the text. Assessing human annotation against high-quality reference labels, we find that annotators frequently misinterpret political content. However, informing the annotators about the political affiliation of the authors yields improved stance annotation, leads annotators to identify more tweets as potentially sarcastic. Overall, 44\% of the tweets in our datasets are perceived as possibly or definitely sarcastic. 

We evaluate several LLMs on our dataset to investigate whether and how they can infer and leverage author context to improve their understanding of political texts. Our results using pre-trained LLMs show that they underperform human annotators on both author affiliation prediction and stance detection. They also struggle more than human annotators to interpret texts labeled as sarcastic. Standard fine-tuning on labeled examples enables the models to outperform human annotators across all categories. However, cross-dataset evaluations show a substantial drop in stance detection performance under distribution shift. This finding is consistent with prior research suggesting that fine-tuned models tend to rely on dataset-specific linguistic cues that may not generalize to new data distributions~\cite{carleyIPM22,xueACL26}.

We observe that providing author context through prompting, indicating whether the author is most likely a Republican or a Democrat, improves the performance of pretrained LLMs, although they leverage it less effectively than human annotators do. Consequently, we introduce a contextual author-informed fine-tuning approach~\cite{choi25}, in which LLMs are trained to predict stance based on both tweet content and the author's likely political affiliation. This approach achieves the best overall performance and improves cross-dataset generalization. Notably, these gains persist when using imperfect political affiliations predicted by the models themselves. Importantly, contextual inference capture also self-critical stances, suggesting that the models leverage both author context and textual content rather than relying solely on political affiliation to predict stance.

Overall, this work makes several key contributions. We demonstrate that author context improves political stance interpretation by both human annotators and LLMs. Reframing stance detection as an author-informed task highlights the complementary roles of political affiliation and target-specific stance prediction, enabling better interpretation of sarcasm and within-camp criticism. We further show that fine-tuning LLMs to predict authors' political affiliations and incorporating this information through contextual fine-tuning yields stance detection performance that surpasses human performance while improving cross-dataset generalization. Finally, we release our dataset of political tweets with author context, to support further research. More broadly, we hope this work facilitates more accurate and author-aware approaches to social text understanding with LLMs.

\section{Related work}
\label{sec:related}

Political stance detection has received sustained research attention over the past several decades. Our findings align with prior studies documenting low inter-annotator agreement in this task~\cite{josephEMNLP17,shenEACL21,carleyIPM22,hovyACL25}, showing that pairwise agreement among low-confidence annotators can fall below chance levels~\cite{josephEMNLP21}. various factors have been proposed to account for the challenges of human stance annotation, including the complexity of stance targets, annotators' levels of political engagement and familiarity with the social media platform, and socio-cultural differences between authors and their audiences~\cite{pexman2005,allawayEMNLP20,shenEACL21,belayTACL26}. Our work extends this line of research by highlighting and examining the contribution of author context to stance annotation, particularly for ambiguous or sarcastic texts~\cite{opreaACL19}.

Computational approaches to stance detection typically leverage sentiment-related features, topical information, and lexical choices~\cite{aldayelIPM21,allawayEMNLP20}. Author-level information has been incorporated into traditional models, primarily through network features that capture homophily among connected individuals in social graphs~\cite{yangTACL17,darwishAAAI20,jiangWSDM23,villaPLOS25}. With the growing adoption of LLMs for stance detection and labeling~\cite{zhangICLR24,gul24,CruickshankTIST25,pendzelEMNLP25,geraSem26}, effectively integrating social context with textual evidence remains an open challenge~\cite{soniACL24,hovyACL25}. For instance, recent efforts to augment LLMs with contextual information retrieved from Wikipedia and web searches have yielded mixed results and have not consistently improved stance detection performance~\cite{pujariACL24,nguyenACL25}.

Unlike related  works on political stance detection~\cite{rostamiEMNLP25,sucuACL25}, we distinguish individuals' stances toward specific targets from their broader political affiliations, allowing us to capture intended sarcasm and within-camp criticism. In doing so, we make a step towards bridging political stance detection with approaches that predict stances based on sociodemographic attributes~\cite{heinischNAACL24}, or leverage such attributes for cross-target stance detection~\cite{allawayEMNLP20,pendzelEMNLP25}. Finally, unlike existing resources~\cite{liACL21,kawintiranonACL21,pujariEMNLP24}, our larger dataset integrates author-level information, author-informed stance annotations, and human judgments of perceived sarcasm, enabling the development and evaluation of models for author-informed political stance detection.

\section{Methodology}

\subsection{Task formulation}
We extend the standard formulation of the stance detection task, in which the objective is to determine from a given text whether the author's is {\it in favor}, {\it against}, or {\it neutral} towards a specified target~\cite{semeval16,liACL21}. Re-framing this task as {\it author-informed} stance detection, and focusing on political tweets, we also represent the political leaning of the text author in categorical form, distinguishing between the classes of {\it Republican} (conservative) and {\it Democrat} (liberal)~\cite{shenEACL21}. 

As we argue and show, acknowledging the political affiliation of the author is often necessary for interpreting their stance from individual tweets. 
Importantly, while one's stances often align with their political leaning on disputable topics, individuals may also criticize their own party and leaders. The extended multi-dimensional stance annotation scheme enables to account for these nuances. We will address the main following research questions:
\begin{itemize}
\item To what extent does providing information about an author's likely political affiliation improve stance annotation and detection from individual texts by human annotators and pretrained LLMs?
\item To what extent can humans and pre-trained LLMs identify the political affiliation of the author and their stances from individual texts? 
\item How can LLMs be adapted to infer and leverage authors’ political learning to improve the accuracy and generalizability of political stance detection?
\end{itemize}

To address these questions, we construct a dedicated dataset of political tweets annotated for stance and supplemented with high-quality author information (\S~\ref{sec:sampling}), having the labels of test examples verified for evaluation purposes (\S~\ref{sec:test}). 

\subsection{Dataset construction}
\label{sec:sampling}

To support this research, we created and annotated a large-scale dataset of political tweets that includes the authors’ political affiliations.

\paragraph{Political affiliation labels.}
It is established that personal attributes, including political affiliation, can be inferred from users’ public social media traces. Specifically, political leanings have been predicted using user networks, textual content, and metadata~\cite{culottaJAIR16,rostamiEMNLP25,rahimzadehACL26}. Otherwise, political learning may be self-declared on some platforms~\cite{sucuACL25}. We build on prior work that maps social media users into a low-dimensional representation space learned from large-scale network data~\cite{elkishkyKDD22,lotanPLOS23}. Specifically,~\citet{lotanPLOS23} learned embeddings for 200K popular Twitter accounts from a large sample of the Twitter network. Treating popular accounts as a `social vocabulary', they represented individual users inductively as a composition of the pre-trained embeddings of the popular accounts that they follow. We use their publicly available pre-trained embeddings to similarly construct social embedding vectors of individual Twitter users based on popular accounts that they follow. Provided with these social user representations, We utilize publicly available compatible classifiers trained in a supervised fashion to predict users’ political affiliations, classifying them as Democrats or Republicans, based on their social embeddings~\cite{lotanPLOS23}.

\paragraph{Political tweet sampling.} To sample political tweets, we drew from a pool of 400K randomly selected U.S. Twitter users, retrieving their most recent tweets and their follow networks as of 2022.\footnote{Tweets were restricted to those posted in 2021 or later.} We identified candidate political tweets by searching for any of the following target terms: {\it Republican(s)}, {\it Trump}, {\it Democrat(s)}, and {\it Biden}. To ensure reliable author information, we retained only tweets whose authors’ political affiliations were predicted from their social networks with high confidence.\footnote{We applied a confidence threshold of 0.93, retaining the 15\% of users classified as either Republican or Democrat with the highest confidence.} This process yielded 22.3K tweets.

When selecting tweets for annotation and evaluation, we aimed to focus on texts that express stance implicitly, posing challenges for automated language processing~\cite{gattoEMNLP23}. To this end, we conducted a preliminary stance annotation step using chain-of-thought reasoning with ChatGPT-4o~\cite{gpt4omini} (\S~\ref{sec:preprocessing}). We prioritized tweets whose predicted stance contradicted the author's political affiliation—for example, tweets critical of Biden or supportive of Trump written by authors identified as Democrats based on their social networks—as such discrepancies could indicate that the intended stance was difficult to interpret. This process identified 3.7K tweets (16.6\%), from which we sampled 1.8K for human annotation, balancing authors from both political camps. To diversify the dataset, we supplemented the sample with 1K tweets whose predicted stance aligned with the author's political affiliation, maintaining a balance between the two political camps. The resulting dataset comprises 2.8K examples, providing a challenging benchmark for training and evaluating author-informed stance detection models.

In a dedicated effort, the authors gauged the predicted political affiliation of 600 authors from this dataset. We considered multiple political tweets authored by each user, relying on unequivocal evidence. Comparing our judgment with the high-confidence network-based predictions showed remarkable accuracy of 96\%. Thus, we consider the network-based assignments to be highly accurate. 

\subsection{A gold-labeled test set}
\label{sec:test}

Recognizing the challenges of annotating individual political tweets, we established high-quality labels for 20\% of the examples to create a reliable test set. To determine the ground-truth stance labels, the paper’s authors examined multiple tweets by each author, including tweets outside our dataset, seeking consistent and unambiguous evidence of their stance toward the target. Disagreements were resolved through discussion. For 3\% of the tweets, insufficient evidence prevented a reliable stance assignment, and these examples were excluded from the test set. We also corrected the network-based political affiliation labels in 2.5\% of cases. Given the subjective nature of sarcasm annotation~\cite{operaACL19} and the fact that its detection was outside the scope of our study, sarcasm labels were obtained exclusively from human annotators.

\begin{table}[t]
\footnotesize
\centering
\begin{tabular}{llccc}
 & & \textbf{CAMP} & \textbf{P-Stance} & \textbf{Elec'20} \\ 
\hline
 & Size &   595 & 205 & 295 \\
\hline
Author & Democrat &	46\%	& 71\%	& 73\% \\ 
     & Republican &	54\%	& 29\%	& 27\%
\\
\hline
Stance: & Favor: Own & 41\% &	26\% & 37\% \\
Target & Against: Other	& 29\%	& 64\% & 48\% \\
 & Against: Own	& 19\%	& 10\% & 15\% \\
 & Neutral &	 9\% &	- & - \\
\hline
Sarcasm & No & 32\% &  &   \\
 & Possibly & 24\% &	\multicolumn{1}{c}{NA}	&  \multicolumn{1}{c}{NA}  \\
& Surely & 20\% &  &  \\
\hline
\end{tabular}
\caption{Test set statistics. The biased sample include tweets that are presumably more challenging.} 
\label{tab:dataset}
\end{table}

Table~\ref{tab:dataset} shows the statistics of the gold labels, which are presumably representative of the full dataset, which we title as CAMP {\it(Context-Aware Modeling of Political stance)}. As shown, 41\% of the test examples represent within-camp support (`favor:own'), 29\% represent cross-camp criticism (`against:other'), and 19\% represent within-camp criticism. This distribution is described in more detail in Table~\ref{tab:dataset-cross} (\S~\ref{app:datasets}). Table~\ref{tab:dataset} also reports statistics for two existing benchmarks, P-Stance~\cite{liACL21} and Elections'20~\cite{kawintiranonACL21}, which we use for cross-dataset evaluation. Neither dataset includes sarcasm annotations, and both likely contain a higher proportion of explicit stance expressions, as discussed in Section~\ref{sec:ft}.

\section{Human judgment}
\label{sec:annotators}

We turned to crowd sourcing, using the Amazon Mechanical Turk (AMT)\footnote{https://www.mturk.com} platform to obtain human judgments of our dataset. First, workers were given the tweets as sole evidence, being asked about: (a) The most likely political affiliation of the author (Republican/Democrat); (b) The prospect of the tweet being ironic or sarcastic, defined as a discrepancy between the literal meaning of the text and its intended meaning~\cite{opreaACL19}
(surely/possibly/no); (c) The perceived stance of the author towards the target specified (in favor/against/neutral). This sequence of questions was meant to raise the workers' attention to the author context and the potential use of sarcasm while annotating the stance. 

In a separate annotation round, we collected {\it author-informed} human judgments by presenting annotators with each tweet alongside the author's likely political affiliation and asking them to answer the remaining questions concerning perceived sarcasm (b) and the author's stance (c).

We recruited experienced, reputable U.S.-based workers who were expected to be familiar with U.S. politics. Before annotating, workers received detailed guidelines and reviewed several labeled examples with accompanying explanations (Appendix~\ref{app:human}). Each human intelligence task (HIT) comprised 10 randomly sampled tweets, and workers received \$1 per HIT. Each tweet was annotated by five workers under each condition, with different workers assigned to annotate the same tweet in the text-only and author-informed conditions.

\begin{figure}[t]
\centering     
\subfigure[Sarcasm]{\label{fig:a}\includegraphics[width=0.49\columnwidth]{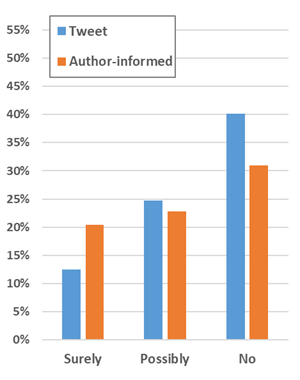}}
\subfigure[Stance]{\label{fig:b}\includegraphics[width=0.49\columnwidth]{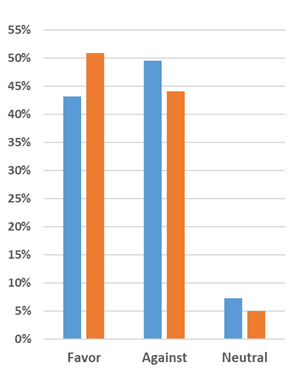}} \\
\caption{Majority labels obtained via crowd sourcing, given tweets only vs. an author-informed labeling setup}
\label{fig:labels}
\end{figure}

\begin{table}[t]
\footnotesize
\centering
\begin{tabularx}{\linewidth}{X}
\hline
(i) “Dark \textbf{Biden} does it again!" \{Dem; favor; Surely\} \\
\hline
(ii) "\textbf{Biden} doesn't have political rallys, he doesn't have worshippers." \{Dem; Favor; No\} \\
\hline
(iii) "Philly needs to elect more \textbf{democrats}. That should fix things." \{Rep; Against; Surely\} \\
\hline
(iv) "\textbf{Democrats} do indeed deliver. All the wrong things. <URL>" \{Dem; Against; Surely\}\\
\hline
(v) "Once again, empty promises from a \textbf{Republican}. Way too many weak men on the right to ever effect change." \{Rep; Against; No\} \\
\hline
\end{tabularx}
\caption{Example human-labeled tweets \{political affiliation, stance, sarcasm\}. The targets are boldfaced.}
\label{tab:examples}
\end{table}

\paragraph{Impact of author context.}

We analyzed human annotations aggregated through majority voting under both the text-only and author-informed conditions. Notably, when annotators were presented with tweets alone, they classified 65\% of the authors as Democrats, despite the dataset's roughly balanced representation of the two political camps. Such misclassifications may have further affected their judgments of perceived sarcasm and stance.

Figure~\ref{fig:labels} compares the distributions of majority-vote labels under the text-only and author-informed conditions. As shown in Figure~\ref{fig:labels}(a), providing author context increased the proportion of tweets labeled as `surely sarcastic' by 8 percentage points, while decreasing the proportion labeled as non-sarcastic by 9 percentage points. These findings suggest that readers tend to interpret tweets literally when relying on textual content alone, whereas author context can help them recognize sarcasm. We also observe a shift in the distribution of stance labels: with author information, the proportion of supportive stances increases by 8 percentage points, while the proportions of critical and neutral stances decrease by 5 and 2 percentage points, respectively (Fig.~\ref{fig:labels}(b)). This shift may reflect a tendency to associate negative sentiment-bearing words with critical stances, whereas author context encourages a more contextual interpretation of such expressions~\cite{allawayEMNLP20}.

To illustrate these shifts, we examine specific examples presented in Table~\ref{tab:examples}. The text (i) "Dark Biden does it again" was initially interpreted as critical of Biden. However, revealing that the author was a Democrat prompted annotators to recognize the sarcasm and revise the inferred stance to supportive of Biden. Similarly, the text (ii), "Biden doesn't have political rallys, he doesn't have worshipers,'' was initially interpreted as criticism of Biden and attributed to a Republican author. Identifying the author as a Democrat led annotators to reinterpret the text as supportive of Biden, despite its negative tone. Example (iii) illustrates the opposite pattern: a seemingly positive statement about {\it Democrats} was initially attributed to a Democratic author. However, revealing that the author was a Republican led annotators to reinterpret the statement as sarcastic and critical of Democrats. Examples (iv) and (v), in contrast, feature strongly negative statements for which author context helped annotators identify within-camp criticism.

\paragraph{Human annotation quality.} Consistent with prior studies, we find that non-expert annotators struggle to infer both a tweet author's political leaning and the stance expressed in the tweet. For political affiliation annotation, Fleiss’ kappa was 0.23, indicating weak inter-annotator agreement~\cite{shenEACL21}. Comparing the majority-vote labels with the gold labels for the test examples, we found that the observed bias toward Democratic affiliation led to higher recall but lower precision for Democrat predictions, with the opposite pattern for Republican predictions. Overall, human annotators achieved a macro-F1 score of 0.67 when inferring tweet authors’ political affiliation.

For stance annotation, we observed a similarly low level of inter-annotator agreement. This may indicate low annotator confidence, reflecting the ambiguity and context dependence of stance judgments, particularly when stance is conveyed implicitly or through sarcasm~\cite{josephEMNLP21}. We evaluated the resulting majority-vote labels against the gold labels for the test examples using weighted F1, given the substantial class imbalance and the relative rarity of neutral stances. When annotators were provided with only the tweet text, the majority-vote labels achieved a weighted F1 score of 0.63. In the author-informed condition, although IAA remained low ($\kappa=0.21$), label quality improved by 18\%, with the weighted F1 score rising to 0.74. This result showcases the role of author context for political text understanding.

Finally, we assessed inter-annotator agreement for perceived sarcasm using Krippendorff’s alpha, treating `possible sarcasm' as an intermediate category between `not sarcastic' and `surely sarcastic'. Again, we observed showed weak agreement, where providing author information increased $\alpha=0.23$ to $\alpha=0.26$. These values compare favorably with prior work reporting low agreement in contextual sarcasm annotation~\cite{abercrombieACL16}. Despite the low agreement rate, which is characteristic of subjective annotation tasks, majority voting across five annotators yields informative labels for subsequent analyses. Given the favorable evaluation results of author-informed annotations, we use these labels to effectively adapt LLMs for political stance detection (\S~\ref{sec:ft}). Using the author-informed sarcasm labels, we investigate and show that perceived sarcasm has a direct impact on both political affiliation and stance detection (\S~\ref{sec:sarcasm}). 

\section{Evaluation of pre-trained LLMs}

In our experiments, we engage with several open-source LLMs, including the models of Llama-3-8B\footnote{https://huggingface.co/meta-llama/Meta-Llama-3-8B}~\cite{llama}, Gemma-7B\footnote{https://huggingface.co/google/gemma-7b-it}~\cite{gemma}, a variant of the Mistral-7B refined via instruction tuning and Direct Preference Optimization~\cite{mistral}, and the proprietary ChatGPT-4o-mini~\cite{gpt4omini}. In applying the pre-trained LLMs to our tasks, we use textual instructions similar to those given to the human annotators (\S~\ref{sec:prompts}). 

\begin{table}[t]
\centering
\small
\begin{tabular}{ll|ll}
 & Affiliation & Stance & Informed \\
\hline
Human & 0.669   & 0.626 & 0.736 \\ 
\hline
\multicolumn{3}{l}{Predict-by-user} & 0.682 \\
\hline
\multicolumn{4}{l}{\textbf{Pre-trained LLMs}} \\
\hline
Mistral-7B	&	\textbf{0.639}	&  0.519   & 0.550 \\
ChatGPT-4o& 0.597 & \textbf{0.527} & \textbf{0.579} \\
Llama3-8B	&	0.479   & 0.475 & 0.522 \\
Gemma-7B	&	0.532	 & 0.342 & 0.265 \\
\hline
\multicolumn{4}{l}{\textbf{Fine-tuned}} \\
\hline
Mistral	&	0.806	& 0.671  & 0.734 \\
ChatGPT-4o& \textbf{0.811} & \textbf{0.778} & \textbf{0.762} \\
Llama	& 0.788	& 0.501 & 0.603 \\
Gemma	& 0.762	& 0.557 & 0.595 \\
\hline
\end{tabular}
\caption{Main results: Political affiliation prediction [Macro-F1] and stance detection [Weighted-F1]} 
\label{tab:main}
\end{table}

Table~\ref{tab:main} reports the performance of pretrained models on the test set.  For political affiliation prediction, Mistral and ChatGPT achieved the highest Macro-F1 scores among the evaluated models, at 0.64 and 0.60, respectively. However, all models fell short of human performance, which reached a Macro-F1 score of 0.67. Detailed results (\S~\ref{sec:app-results}; Table~\ref{tab:author}) reveal a strong bias toward Republican predictions in the Llama and Gemma models, which frequently misclassified Democratic users and consequently achieved low Macro-F1 scores. In contrast, ChatGPT exhibited a bias toward Democratic predictions. These findings align with prior research documenting political biases in LLMs on politically oriented tasks, including a tendency for ChatGPT to exhibit a left-leaning orientation~\cite{fengACL23,motokiPC24,huiWSDM25}.

As detailed in Table~\ref{tab:main}, given the tweets alone, the pre-trained LLMs present poor stance detection performance. ChatGPT performs best among the evaluated models, achieving a weighted F1 score of 0.53, compared with 0.63 achieved by human annotators. To assess the impact of author context, we conducted author-informed predictions by providing the models with information about the tweet author's likely political affiliation alongside the tweet text. As shown in Table~\ref{tab:main}, incorporating author context improved stance detection performance for most models, with gains ranging from 6\% to 10\%. Gemma was an exception: it achieved the lowest overall performance and failed to benefit from the additional author information. These findings suggest that incorporating relevant author context can improve stance detection by pretrained LLMs. 
Nevertheless, these pre-trained models struggle to understand sociopolitical text and perform below human levels under both conditions.

\begin{table}[t]
\centering
\small
\begin{tabular}{llll}
 & Pretrained & Fine-tuned & Informed \\
\hline
\multicolumn{4}{l}{\textbf{Elections'20}} \\
\hline
Mistral-7B	&	0.601 & 0.351 & \textbf{0.752} \\
ChatGPT-4o&	0.662 & 0.499 & \textbf{0.875} \\
\hline
\multicolumn{4}{l}{\textbf{P-Stance}} \\
\hline
Mistral	&	0.815 & 0.726 & \textbf{0.752} \\
ChatGPT-4o& 0.766 & \textbf{0.836} & 0.821  \\
\hline
\end{tabular}
\caption{Cross-dataset stance detection results} 
\label{tab:cross}
\end{table}

\section{LLM fine-tuning}
\label{sec:ft}

Using our dataset, we fine-tuned the LLMs for each task and evaluated their performance on held-out test data. For political affiliation prediction, we used 15K tweets whose authors’ political affiliations were labeled with high confidence based on social network evidence (\S~\ref{sec:sampling}), ensuring no author overlapped between the train and test sets. For stance detection, we used the human-labeled examples. The results of the fine-tuned models are included in Table~\ref{tab:main}. As shown, following fine-tuning, all models achieved Macro-F1 scores of 0.76–0.81 on political affiliation prediction, exceeding the human-level performance of 0.67. Additional experiments varying the number of training examples showed that fine-tuning remained effective with as few as 5K labeled examples (\S~\ref{sec:app-results}; Fig.~\ref{fig:scale}). Similarly, fine-tuning the LLMs on human-labeled tweets for stance detection substantially improved their performance compared to the pre-trained models. ChatGPT and Mistral achieved the highest weighted F1 scores, at 0.78 and 0.67, respectively, outperforming the human-level benchmark of 0.63.

Having established the importance of author context for interpreting ambiguous texts, we conducted author-informed {\it contextual fine-tuning}~\cite{choi25}, prepending a statement about the author’s likely political affiliation to the prompt. This approach aims to enable LLMs to incorporate author-level contextual information when interpreting text. As shown in Table~\ref{tab:main}, the informed setup outperformed standard fine-tuning for most models and achieved comparable performance with ChatGPT, yielding the best overall results for stance detection.

To assess whether author information complements rather than replaces stance detection, Table~\ref{tab:main} also reports results for a baseline that predicts stance solely from the author’s political affiliation. Under this approach, Democratic authors are assumed to express positive stances toward Biden and Democratic targets and negative stances toward Republican targets, with the reverse pattern assumed for Republican authors. Crucially, this baseline cannot capture within-camp criticism or identify neutral stances. As shown in Table~\ref{tab:main}, this ``predict-by-user'' baseline achieves a lower weighted F1 score of 0.68 on the test set. Table~\ref{tab:sc} includes additional results, showing that human workers and LLMs leverage author information without assuming that one's stances always align with their political camp, thus capturing instances of self-criticism.

\paragraph{Cross-dataset evaluation.}

To evaluate the generalizability of the fine-tuned models, we conducted cross-dataset experiments using P-Stance~\cite{liACL21} and Elections'20~\cite{kawintiranonACL21}. Like our dataset, both consist of political tweets collected around 2020 and crowdsourced labels for perceived stance toward Biden or Trump. We retrieved network information for tweet authors from 2022 to infer their political leanings. Since such information was available for only a  subset of users, the resulting evaluation datasets were relatively small, each comprising roughly half the size of our test set (Table~\ref{tab:dataset}). Furthermore, whereas author political affiliations in our dataset were predicted with high classifier confidence (0.97), the corresponding labels in the reference datasets had lower average confidence scores of 0.82 for P-Stance and 0.92 for Elections'20. 

Table~\ref{tab:cross} presents cross-dataset evaluation results for the two best-performing LLMs in our experiments, ChatGPT and Mistral. The pre-trained models achieve weighted F1 scores of up to 0.66 on Elections'20 and 0.82 on P-Stance, compared with 0.58 on our test set. These differences may reflect variations in dataset characteristics: the benchmark datasets may contain more explicit expressions of political stance, as they were collected using hashtags (e.g., \#voteblue). Our dataset also contains a higher proportion of within-camp criticism and neutral stance labels (Table~\ref{tab:dataset}).

Fine-tuning on text alone substantially degrades performance under distributional shift, with weighted F1 scores falling to 0.50 or below on Elections'20. This decline suggests that text-based fine-tuning may encourage reliance on linguistic cues that do not generalize across datasets~\cite{wullachEMNLP21,ludanACL23}. In contrast, author-informed fine-tuning exhibits robustness, raising weighted F1 scores on Elections'20 to 0.88 for ChatGPT and 0.75 for Mistral. On P-Stance, the already strong performance of the pre-trained models is preserved. 

\paragraph{Sensitivity to author context.} Despite the lower confidence of the predicted author affiliations in the reference datasets, incorporating author context improved stance detection performance. In another evaluation, we used the author affiliations predicted for the test examples by the fine-tuned models rather than gold labels. The results of stance detection were only mildly affected: the weighted F1 score decreased from 0.76 to 0.75 for ChatGPT and from 0.73 to 0.70 for Mistral (Table~\ref{tab:main}). These results suggest that the benefits of author-informed contextual fine-tuning largely persist even when author affiliations are inferred rather than known.

\section{The impact of sarcasm}
\label{sec:sarcasm}

\begin{table}[t]
\footnotesize
\centering
\begin{tabular}{lccc}
 & \textbf{No} & \textbf{Possibly} & \textbf{Surely} \\
\hline
\textbf{Human} & 0.709 & 0.615 & 0.616 \\
\hline
\multicolumn{4}{l}{\textbf{Pre-trained}}\\
\hline
Mistral & 0.709 & 0.558 & 0.550 \\
ChatGPT & 0.676 & 0.558 & 0.498 \\
\hline
\multicolumn{4}{l}{\textbf{Fine-tuned}}\\
\hline
Mistral & 0.845 & 0.731 & 0.752 \\
ChatGPT & 0.870 & 0.769 & 0.744 \\
\hline
\end{tabular}
\caption{Author affiliation prediction results by level of perceived sarcasm [Macro-F1]}
\label{tab:sarcasm1}
\end{table}

\begin{table}[t]
\footnotesize
\centering
\begin{tabular}{llll}
 & \multicolumn{1}{c}{\textbf{No}} & \multicolumn{1}{c}{\textbf{Possibly}} & \multicolumn{1}{c}{\textbf{Surely}} \\
\hline
\textbf{Human} &  0.66 / \underline{0.79}$^{\uparrow}$ & 0.60 / \underline{0.67}$^{\uparrow}$ &  0.61 /  \underline{0.75}$^{\uparrow}$ \\
\hline
\multicolumn{4}{l}{\textbf{Pre-trained}}\\
\hline
Mistral & 0.63 / 0.64 & 0.44 / 0.44 & 0.45 / \underline{0.54}$^{\uparrow}$ \\
ChatGPT & 0.64 / 0.68$^{\uparrow}$  & 0.46 / 0.49$^{\uparrow}$  & 0.48 /  \underline{0.58}$^{\uparrow}$  \\
\hline
\multicolumn{4}{l}{\textbf{Fine-tuned}}\\
\hline
Mistral & 0.67 / \underline{0.74}$^{\uparrow}$  & 0.61 /  \underline{0.73}$^{\uparrow}$  & 0.71 / 0.78$^{\uparrow}$  \\
ChatGPT & 0.77 / 0.72$^{\downarrow}$ & 0.76 / 0.77 & 0.81 / 0.79 \\
\hline
\end{tabular}
\caption{Stance detection performance [weighted F1] by sarcasm level for the (text-based / author-informed) setups. Differences exceeding 5\% are indicated by arrows, while those exceeding 10\% are underlined.}
\label{tab:sarcasm2}
\end{table}

We used human-annotated sarcasm labels to examine the effect of sarcasm on political text understanding. Table~\ref{tab:sarcasm1} presents political affiliation prediction performance by perceived sarcasm level. Human annotators were less accurate on sarcastic than on non-sarcastic texts, while pre-trained Mistral and ChatGPT exhibited even larger performance declines. Fine-tuning substantially improved model performance, although accuracy remained lower on sarcastic tweets.

Table~\ref{tab:sarcasm2} reports weighted F1 scores for tweet-based and author-informed stance detection, stratified by perceived sarcasm level. Incorporating author context improves human annotation performance by more than 10\% across all levels, with the largest gain (22\%) for tweets perceived as surely sarcastic. Pre-trained LLMs exhibit a similar improvement ($\sim$20\%) on these tweets. Fine-tuned models are less sensitive to sarcasm, perhaps due to encoding relevant linguistic cues~\cite{davidovCONLL10}. As shown in Table~\ref{tab:cross}, author-informed fine-tuning yields better generalization across datasets.

\section{Conclusion}

Our findings demonstrate that incorporating author context improves political stance detection for both human annotators and LLMs, highlighting its importance for social text understanding. Although pretrained LLMs underperform humans, fine-tuning enables them to predict authors’ political affiliations from language use. Furthermore, author-informed fine-tuning enables stance detection models to surpass human performance and generalize better across data distributions than standard fine-tuning. This research has broad implications, as stance-taking often involves politically contentious targets~\cite{huiWSDM25}. Future work could explore richer representations of authors’ perspectives beyond political affiliations and incorporate author perspectives into model pretraining.

\section{Limitations}

This research relies on datasets and user embeddings derived from Twitter (now X). A key limitation of Twitter data is its limited long-term reproducibility, as accounts may be deleted or suspended and posts removed over time, making data recovery through rehydration necessary. This challenge extends to other social media platforms. Although our study focuses on Twitter data, the proposed approach is generalizable to other platforms where political affiliation labels can be obtained or inferred from social network information or historical posts, including Bluesky~\footnote{\url{https://bsky.social}}~\cite{rostamiEMNLP25} and other social media forums~\cite{rostamiEMNLP25,sucuACL25}. Future work could also extend the evaluation to non-English languages and political contexts beyond the United States. 

Another limitation of our work is its coarse-grained, binary representation of political affiliation, which overlooks ideological diversity within parties, such as the progressive wing of the Democratic Party. Future work could incorporate additional sociodemographic characteristics of stance-takers~\cite{muellerHCI21,pendzelEMNLP25} and move toward more comprehensive representations of author context. Ideally, such representations could be learned through social user embeddings that capture multiple dimensions of an author's background and perspective.

We acknowledge the importance of Other contextual factors in social NLP include conversational context and background knowledge about referenced figures and events~\cite{pujariEMNLP21}. We view author context as orthogonal and complementary to these forms of context, which could be integrated to further improve social text understanding.

\section{Ethical statement}

This research was approved by the Institutional Review Board (IRB; XXX/23). The data used in our experiments was collected from Twitter for research purposes. To protect user privacy, the social network information used to infer authors’ political affiliations was transformed into low-dimensional embeddings derived from aggregated signals based on a subset of popular accounts they follow.

We will make our human-labeled dataset available to researchers. The dataset includes high-confidence political affiliation predictions, under the assumption that user identities cannot be recovered. We may provide access to row data to researchers who are IRB-approved and comply with the same data usage agreement with Twitter (now X). 
Considering that user-level information and utilized for downstream application, as well as the biases that may be inflicted by social user modeling, parties who implement a user modeling approach must inform users of these risks, request and obtain their consent.


\appendix

\section{Tweet sampling}
\label{sec:preprocessing}

We initially employed ChatGPT4o-mini to identify candidate tweets that might be challenging for stance detection by LLMs. We used the following prompt: “Determine if the following political tweet contains a stance with respect to the specified target. If the tweet contains a stance, determine if it is in favor or against and output 'Favor' or 'Against'. Otherwise, if there is no stance in the tweet, output 'No\_stance'. Add an explanation and split with a semicolon.” 
This process generated both a predicted stance and an explanatory rationale for each tweet. We note that despite explicit instructions to restrict outputs to three specified categories,  small number of anomalous outputs were observed. These occurrences were infrequent, comprising fewer than 2\% of the dataset (84 records), and were discarded. 

To estimate the error rate by the model, we manually annotated a small sample of 117 processed outputs. In this analysis, we found 40\% of ChatGPT's predictions as incorrect with high confidence, and additional 23\% as possibly incorrect with lower confidence. In addition to evaluating prediction accuracy, we examined the explanations provided by ChatGPT, as these were explicitly requested during the prediction process. Interestingly, we observed cases in which the explanation appeared sound, yet the final prediction contradicted it and therefore was eventually marked as wrong. To address this inconsistency, we conducted a second round of predictions, as described below. 

\paragraph{Self Refining Chain of Thought.}
Since ChatGPT occasionally output an accurate textual explanation the stance, but still produced incorrect predictions, we employed a Chain of Thought (COT) approach to enhance the prediction process. Concretely, we fed the explanation generated during the first round of annotations back into the model, instructing it to re-evaluate the (non-neutral) stances based on both the original tweet and that explanation. Additionally, we extended the prediction task to include an assessment of the political affiliation of the author, categorizing it as either Republican or Democrat, as implied by the tweet. The prompt used for this round of predictions was as follows: “The following political tweet contains a stance towards the given target. Determine, Using the stance explanation provided to you, whether the specific tweet is in favor or against the target. In addition, determine whether the author of the tweet is a Republican or Democrat. Output first 'Favor' or 'Against' and then 'Democrat' or 'Republican', split by a semicolon.” This iterative process aimed to mitigate prediction errors by leveraging the explanations provided during the first round, thereby enabling a more nuanced and accurate classification.

\paragraph{Post-processing.}
Following the second round of predictions, we eliminated entries where ChatGPT’s reasoning conflicted with its final prediction, removing the records where the model revised its prediction as a result of the COT procedure. This step excluded 444 tweets (approximately 10\%). Additional filters were applied to eliminate near-duplicate tweets, short tweets (with less than 5 words, excluding user mentions or hashtags), and tweets where the target appeared only as part of a mention or hashtag. Following these steps, we were left with a dataset of 2.8K records.  Finally, we selected the records with the top classification confidence scores within each political affiliation (Democrat or Republican), constructing a politically balanced dataset. 

\section{The human annotation process}
\label{app:human}

\paragraph{Worker selection criteria.} The human annotators were required to have completed more than 15K HITs with a historical acceptance rate above 98\% and to hold Master Qualification status. As described in Section~\ref{sec:annotators}, they received fair compensation. Annotators were informed that the dataset included political social media content that could be offensive, and were provided with representative examples as part of the annotation guidelines.

\begin{figure}[t]
\centering
\includegraphics[width=0.95\columnwidth]{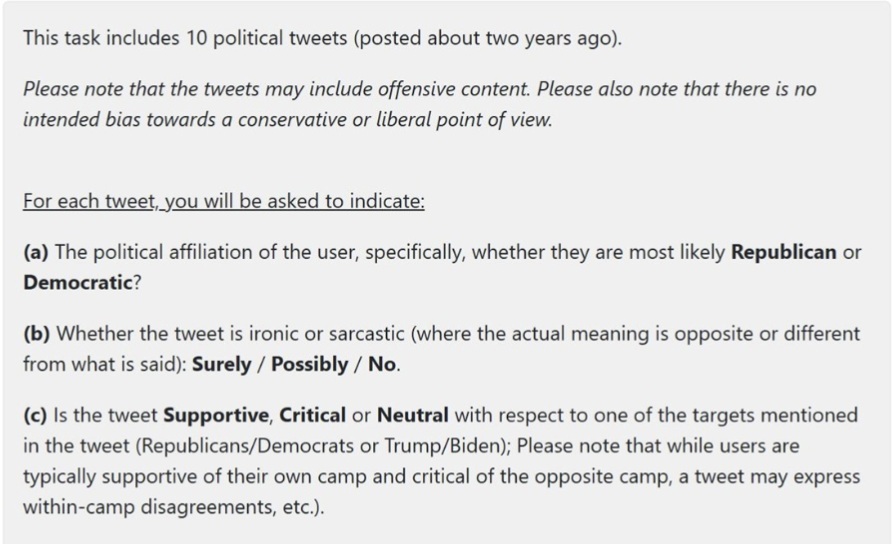}
\small
\caption{The instructions to the annotators}
\label{fig:AMT_instructions}
\end{figure}

\begin{figure}[t]
\centering
\includegraphics[width=0.95\columnwidth]{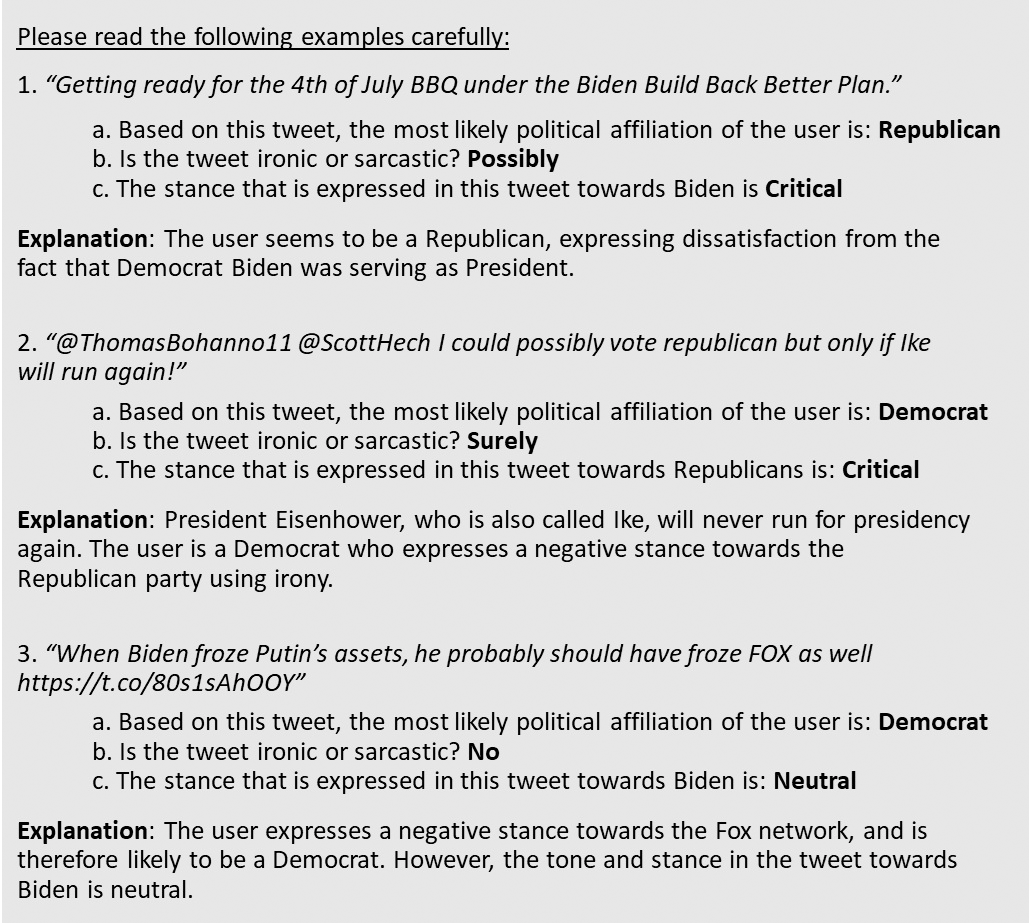}
\small
\caption{Detailed examples presented to the annotators}
\label{fig:AMT-examples}
\end{figure}

\begin{figure}[t]
\centering
\includegraphics[width=0.95\columnwidth]{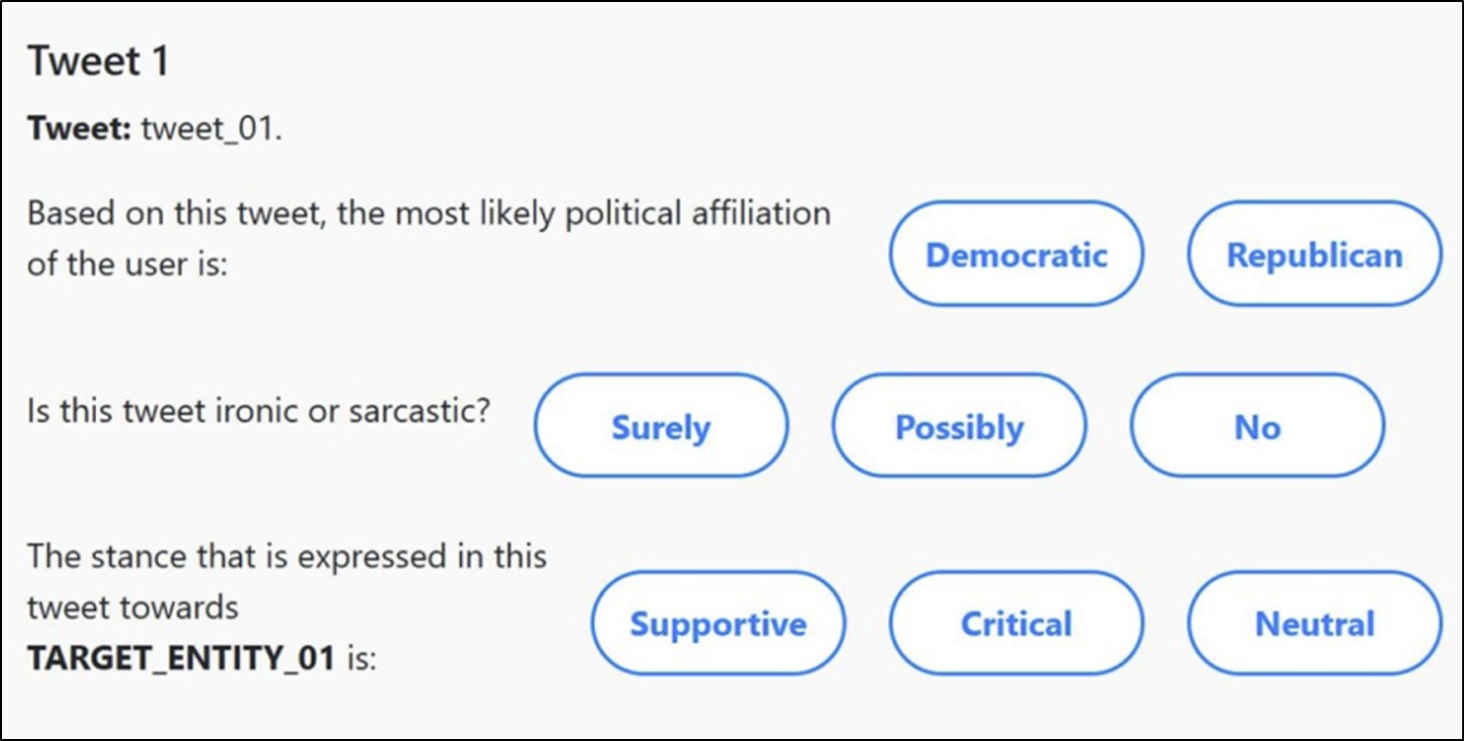}
\small
\caption{The annotation form}
\label{fig:AMT_question}
\end{figure}

\paragraph{Annotator guidance.} Figure~\ref{fig:AMT_instructions} shows our instructions to the annotators. These instructions were followed by examples of labeled tweets along with an explanation for each example, shown in Figure~\ref{fig:AMT-examples}. The annotation interface is shown in Figure~\ref{fig:AMT_question}. As shown, the workers were asked to assign labels in a structured fashion, starting with the presumed author's affiliation, proceeding to assess sarcasm and finally, the stance towards the target term that appeared in the tweet, and was specified accordingly in the annotation form. This approach was intended to encourage the annotators to consider nuanced contextual factors in determining the stance label. In the author-informed setup, we used the same form, stating first the likely political affiliation of the author as Democrat or Republican.

\section{The experimental datasets}
\label{app:datasets}

\begin{table}[t]
\footnotesize
\centering
\begin{tabular}{lllrrrr}
\multicolumn{3}{c}{} & \multicolumn{2}{c}{Test-all} &
\multicolumn{2}{c}{Test-conflict} \\
\hline
User & Stance & Target & \# & Ratio & \# & Ratio \\
\hline
Dem	&	Favor	&	Dem	&	95	&	16.5\%	&	87	&	24.0\%	\\
	&	Neutral	&		&	15	&	2.6\%	&	12	&	3.3\%	\\
	&	Against	&		&	53	&	9.2\%	&	51	&	14.0\%	\\
    \hline
Dem	&	Favor	&	Rep	&	4	&	0.7\%	&	2	&	0.6\%	\\
	&	Neutral	&		&	8	&	1.4\%	&	1	&	0.3\%	\\
	&	Against	&		&	84	&	14.6\%	&	12	&	3.3\%	\\
\hline
Rep	&	Favor	&	Rep	&	141	&	24.5\%	&	119	&	32.8\%	\\
	&	Neutral	&		&	23	&	4.0\%	&	14	&	3.9\%	\\
	&	Against	&		&	55	&	9.6\%	&	53	&	14.6\%	\\
\hline
Rep	&	Favor	&	Dem	&	0	&	0.0\%	&	0	&	0.0\%	\\
	&	Neutral	&		&	4	&	0.7\%	&	0	&	0.0\%	\\
	&	Against	&		&	93	&	16.2\%	&	12	&	3.3\%	\\
\hline
\end{tabular}
\caption{Test set: Distribution of author-stance-target}
\label{tab:dataset-cross}
\end{table}

Table~\ref{tab:dataset} describes the distribution of the labels within the test set. Table~\ref{tab:dataset-cross} further details the distribution of stances, noting the political affiliation of the author and the target entity. This resolution allows to observe the proportion of within-camp and cross-camp support/criticism. We observe that most tweets express within-camp support (e.g., 24.5\% of the test tweets are by Republicans supporting Republicans or Trump) and cross-camp opposition (e.g., 16.2\% of the test tweets are by Republicans criticizing Democrats or Biden). However, some of the tweets express within-camp criticism (e.g., 9.6\% of the test tweets are by Republicans criticizing Republicans or Trump). We also observe that cross-camp support is negligible (e.g., there are no tweets in the test set of Republicans showing support of Democrats or Biden). We note that these statistics are not representative of public opinions, as the tweet sampling procedure was biased. The table details also the tweet distribution within the tweets on which ChatGPT-4O-mini seemed to err in our preliminary assessment (`Test-conflict'). Comparing the two distributions, it seems that the LLMs tended to err on tweets that expressed within-camp criticism. 

\section{LLMs' fine-tuning: implementation details}

The open-weight LLM fine-tuning experiments were implemented using the Hugging Face ecosystem. We employed a QLoRA-style fine-tuning pipeline, loading the base model as a causal language model in 4-bit precision with NF4 quantization and bfloat16 computation. LoRA adapters were applied via PEFT, training only a small set of adapter parameters while keeping the base model frozen. The LoRA configuration used rank 8, alpha 16, dropout 0.05, and targeted the attention projection layers q\_proj and v\_proj. We used Transformers for tokenizer and model loading, Datasets for training data preparation, PEFT for adapter-based fine-tuning, bitsandbytes for 4-bit quantized loading, PyTorch as the backend, and Weights \& Biases for experiment tracking. Training was conducted using the Hugging Face Trainer API with 3 epochs, a learning rate of 2e-4, per-device batch size of 2, gradient accumulation over 8 steps, bfloat16 precision, checkpoint saving every 500 steps, and logging every 10 steps. The model was trained using a causal language modeling objective, where it learned to generate the target label as text rather than being trained as a conventional classifier.

The GPT-based experiments differed in that the model was accessed via OpenAI’s hosted API rather than being loaded locally. For inference, prompts were sent to gpt-4o-mini using system and user messages. For fine-tuning, we used OpenAI’s managed fine-tuning workflow, where formatted training files were uploaded and jobs were executed remotely. No explicit decoding parameters such as temperature, top-p, maximum output length, or penalties were set, so default API settings were used. Each training instance was formatted as an instruction-following example for stance detection, where the prompt contained the task instruction and input (e.g., target and tweet text), and the completion consisted of the gold stance label. The same instruction template was applied across all examples, with only the label varying, enabling the model to learn to generate the correct stance label given the input. Implementation details include the OpenAI SDK for interacting with the Chat Completions API, tiktoken for token counting and cost estimation, and OpenAI’s fine-tuning pipeline for training. Training and validation datasets were uploaded to OpenAI, and the fine-tuning job was run using gpt-4o-mini-2024-07-18 as the base model.

The running times of the fine-tuning experiments were up to three hours using Google Collab (Envidia A-100 GPU).

\section{LLM prompts}
\label{sec:prompts}

This section details the prompts used in our experiments. We have experiments with multiple prompts, and found the below prompts to produce the best and most stable responses overall. A related work examined several prompting schemes, investigating biases of LLMs across various politically oriented stance classification tasks. Their findings suggest that LLMs exhibit varying performance when classifying stance~\cite{huiWSDM25}. They experimented with a 2-phase COT, where the LLM is first instructed to reason about the potential political bias within a statement, and then perform stance classification. However, this strategy was found to exacerbate the observed biases. Unlike their approach, we conducted both tweet-based and author-informed experiments, where the LLMs were provided with external information about the political bias of the tweet author.

The prompt used to predict the author's political affiliation from the tweet is: {\it "Determine the author's political affiliation based on the following political tweet. Classify the author as exactly one of: "Democrat" or "Republican". Output exactly one label, with no explanation or additional text."}

The prompt used in the tweet-based stance detection setting is: 

{\it "Classify the following political tweet with respect to the stance toward the specified target as exactly one of: "Favor", "Against", or "Neutral".
In addition, classify the author's political affiliation as exactly one of: "Democrat" or "Republican". Third, provide a short explanation for both classifications. Before producing the final answer, double-check the tweet stance and author affiliation and revise them if needed. Output exactly three fields separated by semicolons:}
\texttt{TWEET\_STANCE; AUTHOR\_AFFILIATION; EXPLANATION}". 

The prompt used in the author-informed stance detection setting is: 

{\it Classify the following political tweet with respect to the specified target. You are given the likely author’s political orientation, either "Democrat" or "Republican". Use this information as context for interpreting the tweet’s stance.
Classify the tweet's stance toward the target as exactly one of: "Favor", "Against", or " Neutral". Then provide a short explanation for the stance classification. Before producing the final answer, double-check the tweet stance and explanation and revise them if needed.
Output exactly two fields separated by a semicolon:}
\texttt{TWEET\_STANCE; EXPLANATION"}.

\section{Additional results}
\label{sec:app-results}

\subsection{Political stance prediction}
Table~\ref{tab:author} includes detailed results of political affiliation prediction within- and across-datasets. 

\begin{table*}[t]
\centering
\small
\begin{tabularx}{\textwidth}{lccc|ccc|ll|lrlr}
 & \multicolumn{3}{c|}{Democrat} & \multicolumn{3}{c|}{Republican} & \multicolumn{2}{c|}{Within-dataset} & \multicolumn{4}{c}{Cross-dataset: Macro-F1} \\
 	&	Prec.	&	Recall	&	F1	&	Prec.	&	Recall	&	F1	&	\multicolumn{2}{c|}{Macro-F1} & \multicolumn{2}{c}{Elections'20} & \multicolumn{2}{c}{P-Stance'20} \\
\hline
\textbf{Human} & 0.598	& 0.852	& 0.703 & 0.809	& 0.522	& 0.634 & 0.669 &  & \multicolumn{1}{c}{-} & & \multicolumn{1}{c}{-} & \\
\hline
\multicolumn{13}{l}{\textbf{Pre-trained LLMs}} \\
\hline
Mistral-7B	&	0.631	&	0.542	&	0.583	&	0.658	&	0.735	&	0.695	&	\textbf{0.639}	&  & 0.817 &  & 0.679 & \\
Llama3-8B	&	0.776	&	0.140	&	0.238	&	0.574	&	0.966	&	0.720	&	0.479 &  & 0.526 & & 0.487 & \\
Gemma-7B	&	0.525	&	0.343	&	0.415	&	0.575	&	0.742	&	0.648	&	0.532	& & 0.742 & & 0.659 & \\
ChatGPT-4o&	0.546	&	0.679	&	0.605	&	0.664	&	0.529	&	0.589	&	0.597	& & \textbf{0.889} & & \textbf{0.784} & \\
\hline
\multicolumn{13}{l}{\textbf{Fine-tuned LLMs}} \\
\hline
Mistral	&	0.773	&	0.779	&	0.776	&	0.814	&	0.809	&	0.812	&	0.806	&  26\% & \textbf{0.893} & 9\% & \textbf{0.800} & 18\% \\
Llama	&	0.749	&	0.727	&	0.738	&	0.778	&	0.797	&	0.787	&	0.788	&  65\% & 0.889 & 69\% & 0.772 & 16\%\\
Gemma	&	0.697	&	0.712	&	0.704	&	0.755	&	0.742	&	0.748	&	0.762	& 43\% & 0.804 & 8\% & 0.735 & 12\% \\
ChatGPT-40&	0.806	&	0.768	&	0.786	&	0.814	&	0.846	&	0.830	&	\textbf{0.811}	&  36\% & 0.892 & 0\% & 0.778 & -1\% \\
\hline
\end{tabularx}
\caption{Test results: Political stance detection of the tweet author, based on human judgment, and using pre-trained and fine-tuned LLMs. In fine-tuning, we utilized 18K users labeled based on network information with high confidence. Our test examples were manually curated, where the Elections and P-Stance labels are based solely on network evidence.}
\label{tab:author}
\end{table*}

\begin{figure}[t]
\centering
\includegraphics[width=0.95\columnwidth]{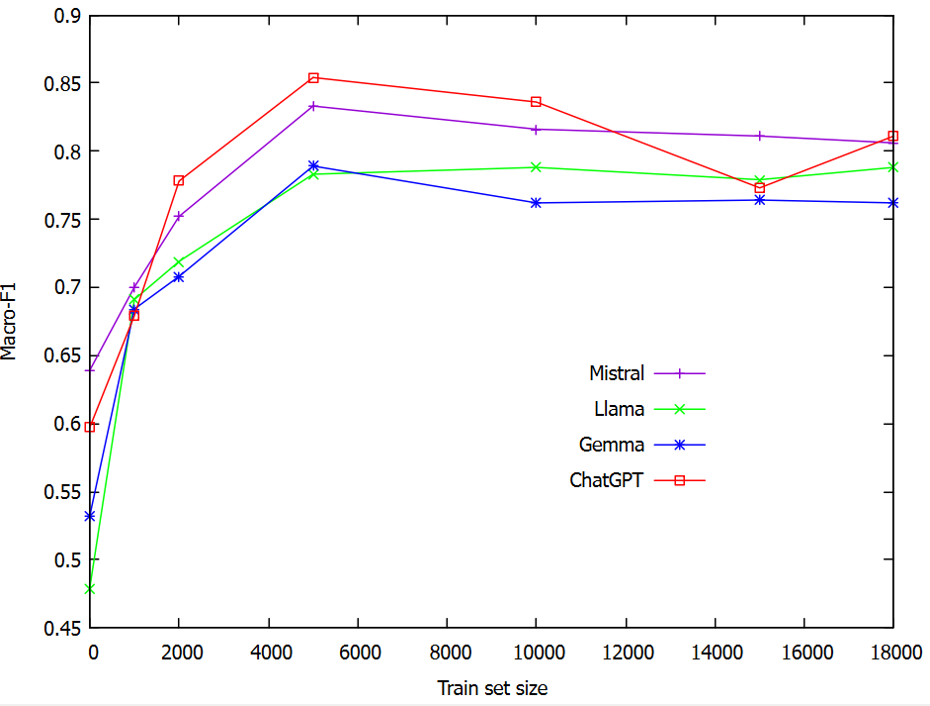}
\small
\caption{Fine-tuned models' test results on political affiliation prediction: Varying train set size}
\label{fig:scale}
\end{figure}

We examined the impact of the train set size on fine-tuning the various LLM models on the task of political affiliation prediction. As shown in Figure~\ref{fig:scale}, while increasing the train set size generally improves the models' test results, we observe that as few as 5K examples support effective fine-tuning. The decline in test set performance past the first 5K examples may be due to prioritizing the train set examples, sampled from the same distribution of the test set, in feeding the examples. 

\subsection{Stance detection}

Table~\ref{tab:sc} presents additional stance detection results, comparing predicted stance–political affiliation combinations against reference labels. We distinguish between four categories: \{self-support, self-criticism, cross-criticism, other\}. The results show that, even after author-informed fine-tuning, models continue to predict stances that contradict the author's political affiliation. This suggests that, as intended, models attend to author context while still attending to textual content. Otherwise, if models had learned to rely primarily on political affiliation as evidence for stance prediction, the proportion of self-criticism predictions would have plummeted.

\begin{table*}[h]
\footnotesize
\centering
\begin{tabular}{llccc}
 & & \textbf{Self-critical} & \textbf{Self-support} & \textbf{Cross-critical} \\
\hline
& \# examples	&	110	&	202	&	129	\\
\hline
ChatGPT & Fine-tuned, author-informed 	&	0.612	&	\textbf{0.820}	&	\textbf{0.839}	\\
	& Fine-tuned, textual &	\textbf{0.651}	&	0.818	&	0.826	\\
    & Pre-trained 	&	0.560	&	0.352	&	0.805	\\
\hline
Mistral	&	Fine-tuned, author-informed 		 & \textbf{0.588}	&	\textbf{0.789}	&	\textbf{0.845}	\\
	& Fine-tuned, textual &	0.573	&	0.773	&	0.748	\\
	&	Pre-trained & 0.545	&	0.416	&	0.806	\\
\hline
\end{tabular}
\caption{Stance detection results, comparing the resulting combinations of predicted stance and political affiliation of the author with the gold annotations, where the label combinations are \{self-support, self-criticism, cross-criticism, or other\}}. 
\label{tab:sc}
\end{table*}

\end{document}